\documentclass[sigconf]{acmart}

\AtBeginDocument{%
  \providecommand\BibTeX{{%
    \normalfont B\kern-0.5em{\scshape i\kern-0.25em b}\kern-0.8em\TeX}}}

\setcopyright{acmcopyright}
\copyrightyear{2024}
\acmYear{2024}
\acmConference[ICCA '24]{ICCA '24: ACM Lemon and Orange Disease Classification using CNN-Extracted Features and ML Classifier}{September 05--06, 2024}{ICCA, Dhaka}
\acmBooktitle{ICCA '24: Lemon and Orange Disease Classification using CNN-Extracted Features and ML Classifier,
September 05--06, 2024, ICCA, Dhaka}
\PassOptionsToPackage{prologue,table}{xcolor}
\usepackage{placeins}
\usepackage[utf8]{inputenc}
\usepackage{algorithm}
\usepackage{algpseudocode}
\usepackage{xcolor}
\usepackage{soul}
\usepackage{fancyhdr}
\begin{document}

%\title{Lemon and Orange Disease Classification using CNN-Extracted Features and Machine Learning Classifier}
\title[Lemon and Orange Disease Classification using CNN-Extractor and ML Classifier]{Lemon and Orange Disease Classification using CNN-Extracted Features and Machine Learning Classifier}

%
%\titlerunning{Lemon and Orange Disease Classification using CNN and ML Classifier}
%\shorttitle{{Lemon and Orange Disease Classification using CNN and ML Classifier}
%\renewcommand{\shorttitle}{}

 \author{Khandoker Nosiba Arifin}
 \authornote{Both authors contributed equally to this research.}
 \affiliation{%
   \institution{Jahangirnagar University}
   %\streetaddress{1 Th{\o}rv{\"a}ld Circle}
   \city{Dhaka}
   \country{Bangladesh}}
 \email{nosiba.stu2018@juniv.edu}

 \author{Sayma Akter Rupa}
 \authornotemark[1]
 \affiliation{%
 	\institution{Jahangirnagar University}
 	%\streetaddress{1 Th{\o}rv{\"a}ld Circle}
 	\city{Dhaka}
 	\country{Bangladesh}}
 \email{rupa.stu2018@juniv.edu}

 \author{Md Musfique Anwar}
 \affiliation{%
   \institution{Jahangirnagar University}
   %\streetaddress{8600 Datapoint Drive}
   \city{Dhaka}
   \state{Bangladesh}}
   %\postcode{78229}}
 \email{manwar@juniv.edu}

 \author{Israt Jahan}
 \affiliation{%
   \institution{Jahangirnagar University}
   %\streetaddress{8600 Datapoint Drive}
   \city{Dhaka}
   \state{Bangladesh}}
   %\postcode{78229}}
 \email{isratju2@gmail.com}

 \renewcommand{\shortauthors}{Nosiba and Rupa, et al.}

\begin{abstract}
  % A clear and well-documented \LaTeX\ document is presented as an
  % article formatted for publication by ACM in a conference proceedings
  % or journal publication. Based on the ``acmart'' document class, this
  % article presents and explains many of the common variations, as well
  % as many of the formatting elements an author may use in the
  % preparation of the documentation of their work.
Lemons and oranges, both are the most economically significant citrus fruits globally. The production of lemons and oranges is severely affected due to diseases in its growth stages. Fruit quality has degraded due to the presence of flaws. Thus, it is necessary to diagnose the disease accurately so that we can avoid major loss of lemons and oranges. 
% To address this issue, we proposed a model that would significantly improve citrus farming by classifying diseases in lemons and oranges.
To improve citrus farming, we proposed a disease classification approach for lemons and oranges.
This approach would enable early disease detection and intervention, reduce yield losses, and optimize resource allocation. For the initial modeling of disease classification, the research uses innovative deep learning architectures such as VGG16, VGG19 and ResNet50. In addition, for achieving better accuracy, the basic machine learning algorithms used for classification problems include Random Forest, Naive Bayes, K-Nearest Neighbors (KNN) and Logistic Regression. The lemon and orange fruits diseases are classified more accurately (95.0\% for lemon and 99.69\% for orange) by the model. The model's base features were extracted from the ResNet50 pre-trained model and the diseases are classified by the Logistic Regression which beats the performance given by VGG16 and VGG19 for other classifiers. Experimental outcomes show that the proposed model also outperforms existing models in which most of them classified the diseases using the Softmax classifier without using any individual classifiers.
\end{abstract}







\maketitle

\section{Introduction}
% ACM's consolidated article template, introduced in 2017, provides a
% consistent \LaTeX\ style for use across ACM publications, and
% incorporates accessibility and metadata-extraction functionality
% necessary for future Digital Library endeavors. Numerous ACM and
% SIG-specific \LaTeX\ templates have been examined, and their unique
% features incorporated into this single new template.

% If you are new to publishing with ACM, this document is a valuable
% guide to the process of preparing your work for publication. If you
% have published with ACM before, this document provides insight and
% instruction into more recent changes to the article template.

% The ``\verb|acmart|'' document class can be used to prepare articles
% for any ACM publication --- conference or journal, and for any stage
% of publication, from review to final ``camera-ready'' copy, to the
% author's own version, with {\itshape very} few changes to the source.
Citrus fruits, specifically lemons and oranges, are vital commodities in the world's agricultural sector, making substantial contributions to both financial gain and dietary needs. In areas with mild to subtropical climates, such as Mediterranean countries (Spain, Italy, Greece), the United States (California, Florida, Arizona), Brazil, China, India, and parts of North Africa, lemons and oranges grow well. Lemons can be found all year round, with winter and spring being the prime harvest seasons. On the other hand, oranges have specific harvesting seasons: navel oranges are usually picked from late fall to early spring, and valencia oranges from spring to early summer. However, variables like temperature and cultivation methods might affect the precise harvesting periods. However, a number of diseases that can negatively impact marketability, quality, and yield make cultivating these fruits difficult. Effective disease management and crop preservation depend on early diagnosis of disease. Conventional techniques for diagnosing in citrus fruits frequently depend on manual inspection by skilled agronomists, which might be time-consuming, subjective, and susceptible to human error. Moreover, to minimize fruit losses and apply focused management strategies depend on how accurately classifying diseases.\\

Nowadays, machine learning approaches to automate disease identification in agricultural sectors have garnered increasing attention. Our research aims to develop a machine learning-based methodology that is specifically designed to classify diseases of oranges and lemons.
% Our approach utilizes a multi-algorithmic framework, integrating the strengths of various machine learning models to improve the accuracy and robustness of disease classification, in contrast to earlier studies that frequently concentrate on a single algorithm or methodology.
Our approach improves disease classification accuracy and robustness by incorporating the strengths of multiple machine learning models, unlike previous research that depend only on one algorithm or methodology.
In particular, We use the convolutional neural network (CNN) architectures of VGG16, VGG19, and ResNet50 for disease classification, which have achieved excellent results in image recognition applications.
% we use the convolutional neural network (CNN) architectures of VGG16, VGG19 and ResNet50 for disease classification, which have proven to achieve state-of-the-art results in image recognition applications.
Furthermore, to enhance classification accuracy and generalization capacity, we incorporate conventional machine learning techniques, such as Random Forests (RF), K-Nearest Neighbors (KNN), Naive Bayes (NB) and Logistic Regression(LR).\\

The research outcomes have great potential for the fruit industry, providing a reasonable and expandable approach to disease control. Our method has the potential to improve crop durability, optimize resource allocation, and ultimately contribute to the sustainability and profitability of fruit production systems by providing producers with timely and accurate information. In a nut shell, the contributions of this research work are as follows:

\begin{itemize}
  \item Most of the existing approaches applied CNN architecture to directly extract features and then classify the images. We utilized different CNN pretrained models such as VGG16, VGG19, ResNet50 for features extraction only and then applied different classification models for better performance.
  
  \item We conduct comprehensive experiments using real datasets to show the effectiveness of our procedure.
  \item We also compare our proposed model with other existing models to demonstrate its effectiveness.
\end{itemize}

\section{Literature Review}
% Several techniques for detecting and classifying diseases of citrus fruits particularly in lemons and oranges have been proposed by the researchers in the field of machine learning.
Machine learning researchers have developed several strategies for detecting and categorizing diseases in citrus fruits, including lemons and oranges.
% Saha et al. \cite{saha2020orange}  proposed deep learning based approach which is used for orange fruit disease classification. This system follows several steps, such as pre-processing unit, segmentation unit, feature extraction unit, and training. 
Saha et al. \cite{saha2020orange}  proposed deep learning based approach which is used for orange fruit disease classification. This system follows several steps, such as pre-processing, segmentation, feature extraction, and training. 
%Classification, then, gives the result. The dataset includes 68 total images, of which 19 indicate citrus canker, 20 indicate melanose diseases, 20 show healthy oranges, and 9 show brown rot. 
The accuracy of the proposed system's classification 
of orange fruit diseases and healthy oranges is 88.89\% for brown rot, 84.21\% for citrus canker,
100\% for melanose, and 100\% for healthy oranges. 
% Gautam et al. \cite{gautam2021analysis} developed a model to use transfer tearning
% methods to analyze the performance for detection and 
% classification of citrus diseases.
Gautam et al. \cite{gautam2021analysis} developed a model to utilize transfer learning to evaluate the performance of detecting and classifying citrus diseases.
%The dataset is collected from different sources and it includes both the leaf and the fruit images of citrus fruits. 1640 images for leaves and 1280 images for fruits. 
The model that utilized 
pre-trained features from Xception performed superior to VGG16 in 
classifying diseased leaves, with accuracies of 93.9\% and 92.7\%, 
respectively, while the Xception-based model outperformed VGG16 
in detecting citrus fruit diseases, with accuracies of 91.4\% and
83.6\%, respectively.
Qingmao et al. \cite{zeng2020gans} suggested GANs-based data augmentation within deep learning.
The dataset contains 5406 images of citrus leaf which is infected by HLB including 
1458 in early infection stage, 2557 in moderately infection stage and 1391 in severely infection stage. Their proposed approach build six different kind of models for detecting the severity of citrus HLB. Among these models, Inception\_v3 model achieves higher accuarcy of 74.38\%. Again, they tried to adopt DCGANs to assess the effectiveness of GAN-based data augmentation in improving model learning performance.  Next, they trained the new model with the dataset of 14,056 leaf images which outperformed the Inception\_v3 model by about 20\%, with an accuracy of 92.60\%.
% achieved accuracy of 92.60\%, almost 20\% higher than the Inception\_v3 model.
For this reason, to improve the model performance effectively they suggested GANs-based data augmentation.

In \cite{rauf2019citrus}, the authors applied selector algorithms. These alogrithm helps them to extract texture, color, and geometric 
features. Then they select those features with the help of skewness, PCA, and Entropy
 methods. Next, their proposed model performs the classification to classify the diseases. %The dataset istaken from a public source at https://data.mendeley.com/datasets/3f83gxmv57/2. The dataset contains 759 images of healthy and unhealthy citrus fruits and leaves.
Poonam et al. \cite{dhiman2022novel} proposed a model that uses
 a deep neural network (DNN) model trained with transfer learning using VGGNet 
to solve the pressing demand for citrus fruit disease detection, after 
preprocessing a dataset. With 99\% accuracy for low severity, 98\% accuracy for 
high severity, 96\% accuracy for healthy circumstances, and 97\% accuracy for 
medium severity, the study highlights the efficiency of their suggested method 
to detect citrus fruit diseases accurately at various severity levels.
The reseaerch work \cite{khattak2021automatic} suggests an automated detection approach using convolutional neural networks
(CNNs) to solve the crucial issue of reducing citrus fruit yields because of diseases.
Utilizing an integrated technique, they aim to reliably identify citrus fruits and 
leaves that are healthy from those that have been affected with prevalent diseases 
such as canker, scab, black spot, melanose and greening. The studies performed with 
the citrus and plantVillage datasets demonstrate that their CNN model performs better 
than other deep learning models, with an impressive test accuracy of 94.55\%. This indicates that it has the potential to be a valuable tool for farmers who want to effectively classify citrus fruits and leaf diseases.
The research \cite{xing2019citrus} represents a CNN model that is customized for their own image dataset and is more effective than pre-trained models. Their experiment's results demonstrate that the weakly DenseNet-16 performs better at classifying data with less parameters and may be used on mobile devices due to its lightweight construction.

The research work \cite{elaraby2022classification} explains a method that uses deep learning and image processing to identify and classify diseases in citrus plants. Five different citrus disease types, including black spot, anthracnose, canker, scab, greening, and melanose, are included in the experimental dataset. AlexNet and VGG19, two varieties of convolutional neural networks, were employed in the development and assessment of the suggested model. At its peak, the accuracy of the entire system was 94\%. 
For classifying diseases and grading of orange, the authors in \cite{8524415} applied machine learning and fuzzy logic to perform two tasks: 1) Orange disease classification; 2) disease severity assessment. They have gathered four categories of infected orange samples, including those with brown rot, canker, obstinate, and melanose, from the University of California Agriculture and Natural Resources database. Fuzzy logic is used to calculate the disease severity of four different types of diseased oranges, and multi-class SVM with K-Means clustering with 90\% accuracy is introduced for the purpose of disease classification. %But they don't classify the disease severity level such as low, medium and high.
In the early 2000s, canker, a serious citrus disease in Florida, began to affect grapefruit, one of the citrus kinds most susceptible to it. In \cite{yadav2022citrus}, they used  Ruby Red grapefruit to detect the citrus disease. They collected the fruit samples that were considered marketable but had canker, greasy spots, melanose, scabs, insect damage, wind scars, and other frequent diseased peel diseases from the University of Florida. They developed a convolutional neural network(CNN) architecture which is a modified versions of VGG16 and attached the VGG16 to the Softmax classifier for multi-class classification. For comparing the CNN's performance they used five randomly selected PCA bands. Zongshuai Liu et al. \cite{liu2021image} presents a deep learning-based solution for the visual detection of citrus diseases. They used a citrus image dataset consisting of six prevalent citrus diseases, including canker,anthracnose, black spot, sand paper rust, huanglongbing, and scabis. %they collected these images via Internet search and field photography. 
The proposed model applied MobileNetV2 model as a deep learning network and compare it with other pretrained architectures such as ResNet50, DenseNet201 and InceptionV3 with respect to the speed, accuracy, and model size. Among these architectures MobileNetV2 provides the better accuracy which is 87.28\%.
Since plant diseases are rather common, diagnosing diseases in lemon trees is essential in the agricultural sector. In the paper \cite{solanki2022lemon}, they have used different conventional neural networks such as ResNet, GoogleNet, and SqueezeNet models with and without data augmentation. A dataset comprising five classes—blackspot, canker, greening, melanose, and healthy—was gathered from Mendeley data and utilized in this work. 
The ResNet model gives the best accuracy with 97.66\%. In the paper \cite{Dillip2022}, they also used the same dataset and same methods such as ResNet50, GoogleNet and SqueezeNet with 256 × 256 pixels in size, 72 dpi resolution, and dataset splitting as same as \cite{solanki2022lemon}. In this case, among the developed models with the data augmentation the SqueezeNet gives the best performance with the accuracy 99.07\%.

In recent years, the agricultural sector has faced significant challenges due to the impact of diseases on orange fruit production \cite{Goyal2024Comprehensive}. This review paper emphasizes on the importance of early disease detection to reduce the financial losses and raising the quality of oranges. They describes different diseases and their symptoms that might harmful for the orange crops and the difficulties associated with the traditional and manual techniques of detection. With Ravinderjit and others in \cite{Ravinderjit2019Detection} suggested a method for utilizing the SVM and K-Means algorithms to detect and classify orange leaf disease. Utilizing SVM and the HK-Means clustering approach is the main goal. The dataset they employed for detection and classification consisted of an album of photos showing both healthy and diseased plants and leaves, with the diseases canker, blackspot, and melanose identified. A total of 750 photos with a resolution of 72 dpi and 256 × 256 dimensions are included in the dataset. %After pre-processing the images, image segmentation is performed using the boundary and spot detection algorithms. 
They employed the HOG technique to extract features and the K-Means algorithm to segment images. This system has 98\% overall accuracy.

\section{Proposed System}
In this research, we represent a computer vision system for classifying the lemon and orange diseases. Disease classification is one of the most important modules in our research. In this module, our target is to classify the lemon and orange fruit diseases more accurately using different types of approaches. Here, we build three different types of models to evaluate the performances and then based on these performances we make a decision that which model is the best for our datasets. Our system classifies the diseases by extracting some of the external features of the lemon and orange images. The feature extraction is done by the predefined models such as VGG16, VGG19 and ResNet50. The disease is then classified as lemon and orange, respectively, using the classifiers (Naive Bayes, Random Forest, K-Nearest Neighbors, and Logistic Regression). The proposed model's architecture is depicted in Figure 1.
%algorithm hbeee
% \input{algorithms/disease_classification}

\begin{figure}[ht]
  \centering
  \includegraphics[width=\linewidth]{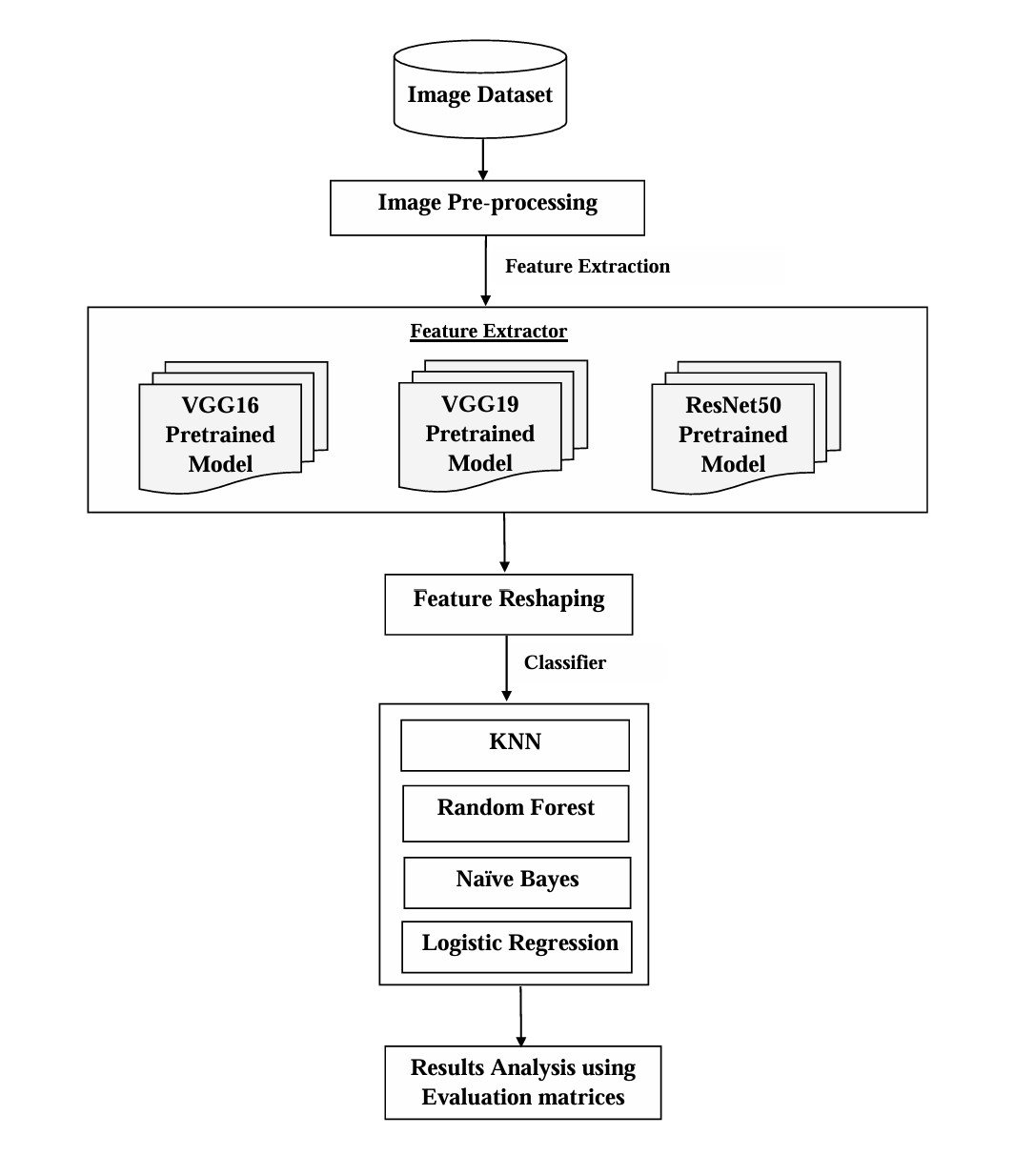}
  \caption{Proposed Model for Disease classification Approach.}
  %\Description{The 1907 Franklin Model D roadster.}
  \Description{This is our proposed methodology}
\end{figure}

% \input{algorithms/disease_classification}

% In \texttt{Disease Classification} (shown in Algorithm \ref{alg:disease_classification}), we first split the dataset after performing some prepossessing steps such as resizing, normalization etc. (line 1-2). Then, we applied different pre-trained models to extract and reshape features (line 3-5). Next, we select different machine learning classifier models to train the model in order to predict class labels for test data samples (line 6-8). Finally, it evaluates and presents experimental outcomes to demonstrate the performance of the proposed strategy (line 9-10).

% \begin{algorithm}
% \caption{\texttt{Disease Classification}}\label{alg:disease_classification}
% \begin{algorithmic}[1]
%     \State Load and pre-process (resize, normalize, etc.) the image dataset.
%     \State Split the dataset.
%     \State Select a pre-trained model as the base model (e.g., VGG16, VGG19, ResNet50).
%     \State Extract the essential features from the base model.
%     \State Reshape the extracted features.
%     \State Select a classification algorithm (e.g., K-Nearest Neighbors (KNN), Random Forest, Naive Bayes, Logistic Regression).
%     \State Train the model using the training dataset.
%     \State Predict the class labels for the test dataset.
%     \State Evaluate the performance of the classifier using accuracy, recall, precision, and f1 score.
%     \State Visualize the experimental results (e.g., confusion matrix, classification report).
% \end{algorithmic}
% \end{algorithm}

\subsection{Image Dataset}
In this study, we used two datasets. The lemon dataset contains 200 images and 1614 images for orange dataset. Both of these datasets are obtained from kaggel but it is noteworthy that these datasets are not directly used by any existing research paper.  

\subsubsection{Dataset Acquisition}
The lemon dataset consists of four classes such as lemon canker, lemon mold, lemon scab, and healthy lemon. On the other hand, orange dataset also consists of four classes and they are blackspot, canker, fresh, and greening. These datasets play an important role in training and evaluating our system model for disease classification of lemons and oranges. The lemon dataset used in this study is from \cite{farwabatool_dataset}  and was obtained via the Kaggle platform. Similarly, the orange dataset in our study is taken form \cite{orange_diseases_dataset} which also got via kaggle.

\subsubsection{Dataset Description}
Both lemon and orange datasets consist of a set of high-resolution images showing both healthy and unhealthy lemons and oranges. Annotations on each image provide details about the particular disease type. The variety of lemon and orange conditions in the dataset guarantees a thorough representation of the several diseases that arise during the cultivation of lemons and oranges. The number of lemon and orange dataset samples is as shown in Tables 1 and 2, respectively.

\begin{table}[ht]
\centering
\caption{Number of lemon dataset samples}
\label{tab:lemon_samples1}
\setlength{\tabcolsep}{12pt}
\begin{tabular}{lc}
\toprule
\textbf{Disease class} & \textbf{Number of images} \\ 
\midrule
%\textbf{Disease class} & \textbf{Number of images} \\ 
Lemon Canker           & 50                        \\ 
Lemon Mold             & 50                        \\ 
Lemon Scab             & 50                        \\ 
Healthy Lemon          & 50                        \\
\bottomrule
\end{tabular}
\end{table}

\begin{table}[ht]
\centering
\caption{Number of orange dataset samples}
\label{tab:orange_samples3}
\setlength{\tabcolsep}{12pt}
\begin{tabular}{lc}
\toprule
\textbf{Disease class} & \textbf{Number of images} \\ 
\midrule
Blackspot              & 344                        \\ 
Canker                 & 349                        \\ 
Fresh                  & 552                        \\ 
Greening               & 369                        \\ 
\bottomrule
\end{tabular}
\end{table}

% 
%eta lemon disease er table
%\begin{table}[ht]
%\centering
%\caption{Number of lemon dataset samples}
%\label{tab:lemon_samples}
%\begin{tabular}{|c|c|}
%\setlength{\tabcolsep}{45pt}
%\begin{tabular}{lc}
%\hline
%\textbf{Disease class} & \textbf{Number of images} \\ \hline
%Lemon Canker           & 50                         \\ 
%Lemon Mold             & 50                         \\ 
%Lemon Scab             & 50                         \\ 
%Healthy Lemon          & 50                         \\ \hline
%\end{tabular}
%\end{table}%
%\vspace{-20pt}
%eta orange disease er table
%\begin{table}[ht]
%\centering
%\caption{Number of orange dataset samples}
%\label{tab:orange_samples}
%\setlength{\tabcolsep}{45pt}
%\begin{tabular}{lc}
%\hline
%\textbf{Disease class} & \textbf{Number of images}   \\ \hline
%blackspot              & 344                         \\ 
%canker                 & 349                         \\ 
%fresh                  & 552                         \\ 
%greening               & 369                         \\ \hline
%\end{tabular}
%\end{table}%

The descriptions of the disease classes are as shown in Figure 2.
\begin{figure}[ht]
  \centering
  \includegraphics[width=\linewidth]{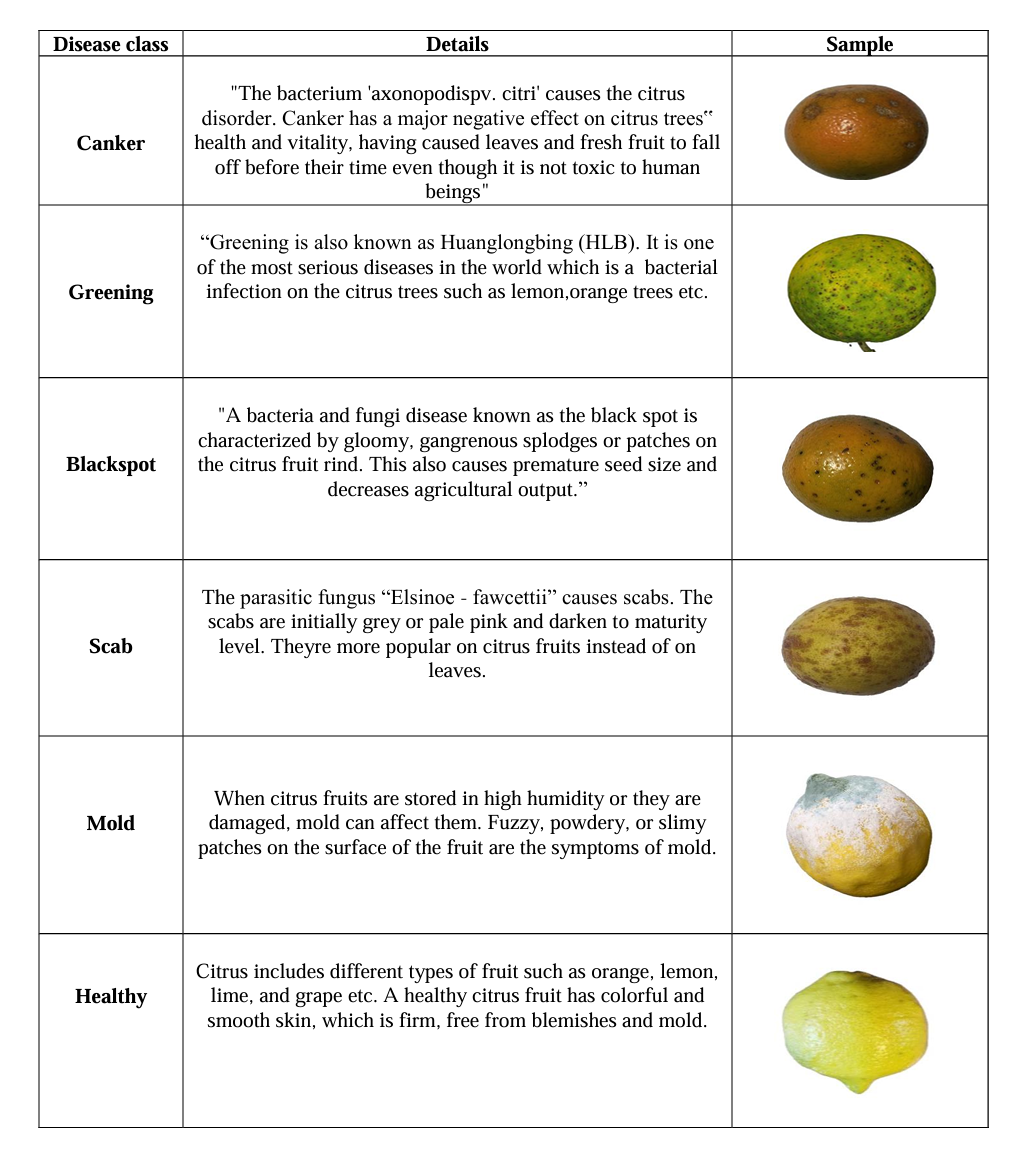}
  \caption{Fruit Disease classes.}
  \Description{This denotes the fruit disease classes}
\end{figure}
\subsection{Image Pre-processing}
In this module, we follow several steps. These are: \textbf{(1) Load and Resize Image:} Images are loaded from the appropriate location and resize them to specific size (224, 224 pixels). Then convert the image from BGR to RGB  to ensure compatibility with further processing stages. Resized images are kept in lists for further analysis, along with the class names that relate to them. \textbf{(2) Split and Visualize Data:} The dataset is splitted into two parts; 
80\% of the entire dataset is put aside for training, while the remaining 20\% is set aside for testing. Randomly selects images from the training dataset and displays them with their corresponding class labels. The lemon and orange images with class labels are visualized in Figures 3 and 4.
\begin{figure}[ht]
  \centering
  \includegraphics[width=\linewidth]{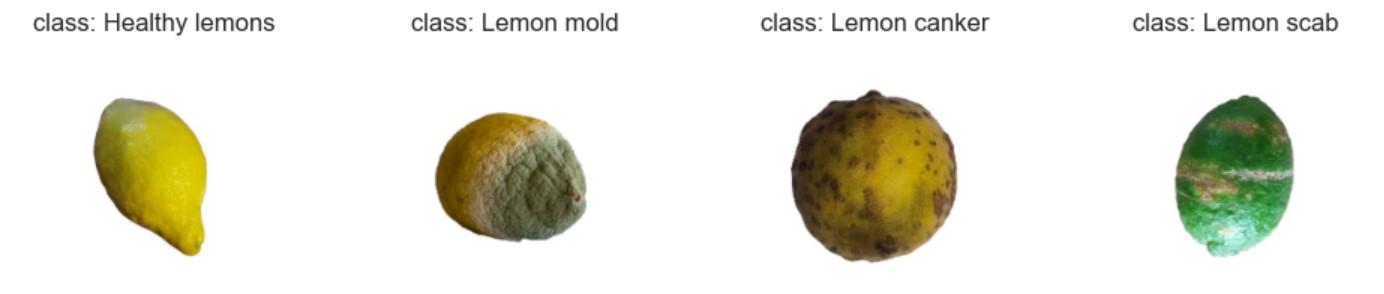}
  \caption{Visualization of lemon images with class label.}
  \Description{This denotes the visualization of lemon images with class label}
\end{figure}

\begin{figure}[ht]
  \centering
  \includegraphics[width=\linewidth]{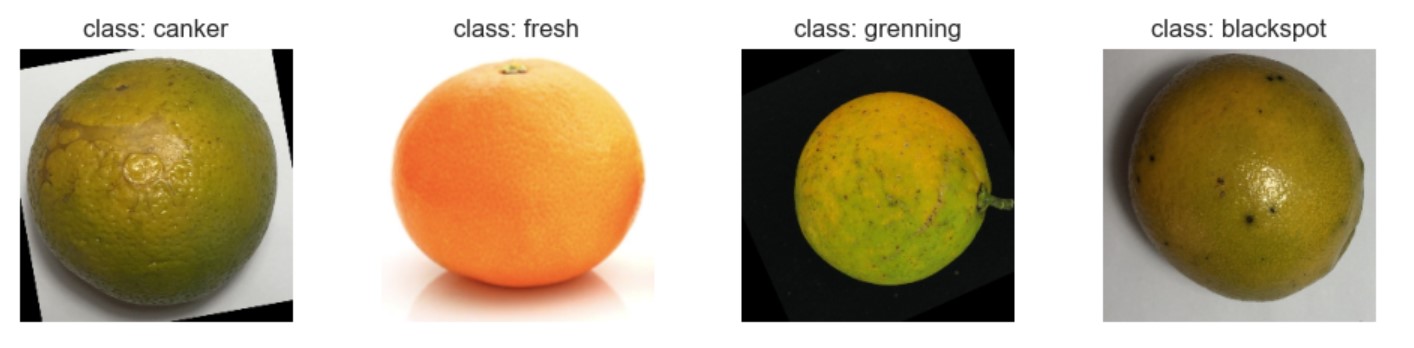}
  \caption{Visualization of orange images with class label.}
   \Description{This denotes the visualization of orange images with class label}
\end{figure}

\subsection{Feature Extraction}
This module involves minimizing the amount of resources needed to accurately describe an enormous quantity of data. One of the major challenges with analyzing complicated data is that it usually involves an excessive amount of memory and processing resources due to the huge number of variables, or it uses an algorithm for classification that overfits the training set and performs poorly when applied to a new sample. In this study, the pretrained CNN architecture VGG16, VGG19 and ResNet50 are selected for feature extraction, allowing for a thorough analysis for disease classification in lemons and oranges. 
\begin{figure}[ht]
  \centering
  \includegraphics[width=\linewidth]{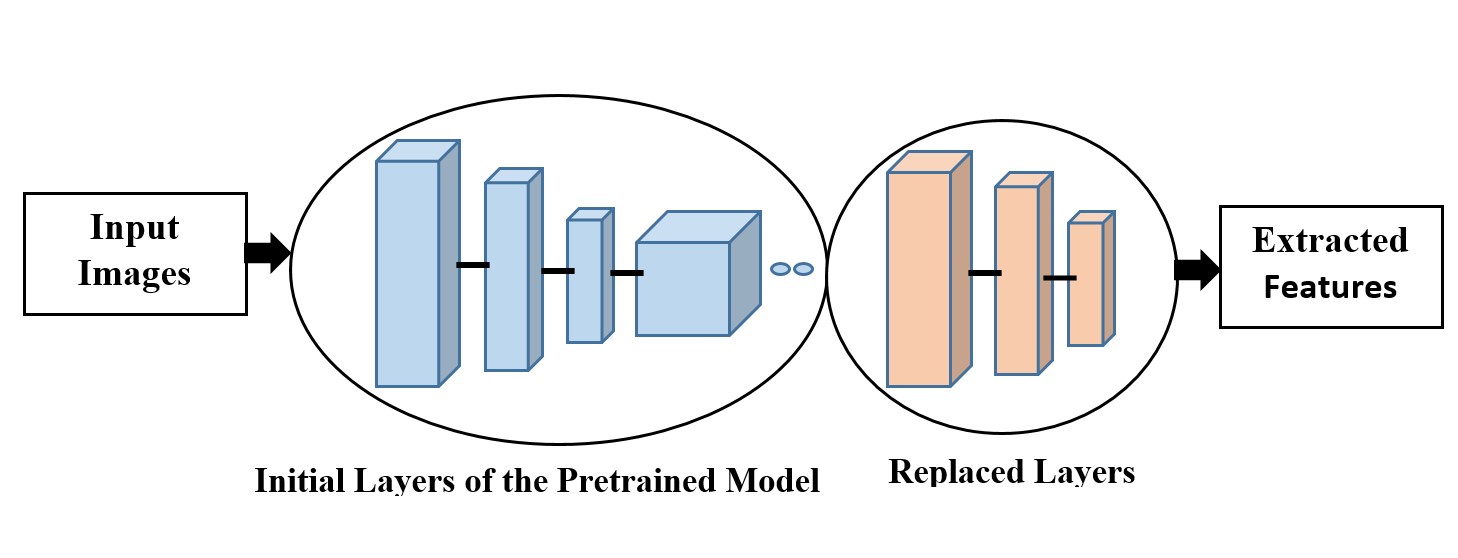}
  \caption{Structure of Pretrained Model.}
  \Description{This denotes the structure of Pretrained Model.}
\end{figure}

These pretrained CNN architectures have been trained on a large image dataset like ImageNet specially for image recognition tasks and improve the model performance on another different dataset. A wide range of features and patterns of images have already been discovered to these models. Pretrained models can be used to start with tasks that change significantly from the first training job, even if they are not the same.
Figure 5 illustrates the structure of these pretrained models. These pretrained architectures are good at extracting features. The features that we have acquired from the network's intermediate layers can be applied to a variety of computer vision applications or used as a basis for other machine learning models. For disease classification, the fully connected layers of the networks are eliminated, and features are created from the last convolutional layer's output. In our study, the feature extraction process doesn’t include any unnecessary features. This is ensured through the following steps: 
\begin{itemize}
    \item we use well-established CNN models (VGG16, VGG19, and ResNet50) that were already pre-trained on the ImageNet dataset, ensuring the extraction of relevant features.
    \item We then apply L1 and L2 regularization techniques in our dense layers to prevent overfitting and ensure that the extracted features are essential and contribute to the model's performance. 
    \item The output features of the CNN models are flattened to make them consistent with the input specifications of conventional machine learning classifiers, retaining key features while eliminating unnecessary complexity.
\end{itemize}
%\sethlcolor{yellow}
By utilizing these techniques, we ensure that the feature extraction process is efficient, focusing only on relevant and essential features for accurate classification.

Every image in our dataset is passed through these networks, and a feature vector is produced from the final convolutional layer's output. This vector represents the high-level properties of the input image that the corresponding CNN learned.

\subsection{Feature Reshaping}
In this research, after extracting the feature from the pre-trained model, the extracted features are reshaped into flattened arrays.  The study's analytical capacities for disease categorization in lemons and oranges are improved by this technique, which makes it easier to integrate feature representations with conventional machine learning algorithms. This reshaped one-dimensional array is suitable for classifier algorithms like KNN, Random Forest, Naive Bayes, and Logistic Regression.
\subsection{Classification}
After reshaping the extracted features from the images using VGG16, VGG19, and ResNet50, we move on to the classification step. We employed four distinct algorithms—K-Nearest Neighbors, Random Forest, Naive Bayes, and Logistic Regression—for the classification of diseases. 
\textbf{K-Nearest Neighbors: } One of the most straightforward and often used supervised machine learning techniques is K-Nearest Neighbors, which classifies a dataset by utilizing the majority label of the nearest neighbor in the feature space. We experimented with several values of k to see how well the K-Nearest Neighbors (KNN) classifier performed.
\textbf{Naive Bayes: } Another algorithm for supervised machine learning is called Naive Bayes which assumes feature independence and is based on the Bayes theorem. Because of its simplicity and speed, it is excellent at predicting the correct class for given features, which makes it useful for multi-dimensional data classification problems. Naive Bayes's main benefit is its rapid classification speed, which is very helpful for large datasets. \textbf{Random Forest: } Random Forest is a popular classifier which builds several decision trees using various dataset subsets and chooses the final prediction by a majority vote among all the trees. It can handle tasks including both regression and classification, making use of its ensemble nature to improve prediction accuracy. This algorithm involves to select random subsets of data to create each tree, then it aggregates their predictions to determine the final outcome. \textbf{Logistic Regression: } Logistic Regression is a fundamental classification algorithm which is a linear model. It is used for binary classification to calculate the likelihood that a sample will fall into a specific class. We also evaluated the algorithm's performance on our extracted features, given its efficacy for large datasets and simplicity of use.

\section{Experimental Evaluation}
 The orange dataset contained 1291 training files and 323 testing files, as we used 80\% of the total dataset for training and 20\% for testing. Similarly, we have 160 training files and 40 testing files for the lemon dataset. In this section, we present and evaluate the outcomes we obtained from our developed models, as well as compare them to existing methods namely VGG16 and ResNet50.
\
 In our proposed model, at first the feature extraction process has occurred. After extracting the features, we can categorize the disease of lemon and orange fruits based on these characteristics. In this case, we created three disease classification models using three different CNN pre-trained architectures, including VGG16, VGG19, and ResNet50. Then, to evaluate how well these models performed, we employed four classifiers: KNN, Naive Bayes, Random Forest, and Logistic Regression. We also evaluated another hybrid method known as ensemble method that under-performs in comparison to our proposed model. Since we used two datasets in this study, one for orange and one for lemon, we applied each model to both datasets to observe their performances. After developing all models, we obtained the highest accuracy for the feature extractor ResNet50 with the classifier named Logistic Regression.
 
 \ For the lemon dataset, Logistic Regression provides 95\% accuracy, 94.10\% recall, 94.32\% precision, and 93.88\% f1 score. The classification of diseases using Logistic Regression is illustrated in Figure 6.

\begin{figure}[ht]
  \centering
  \includegraphics[width=\linewidth]{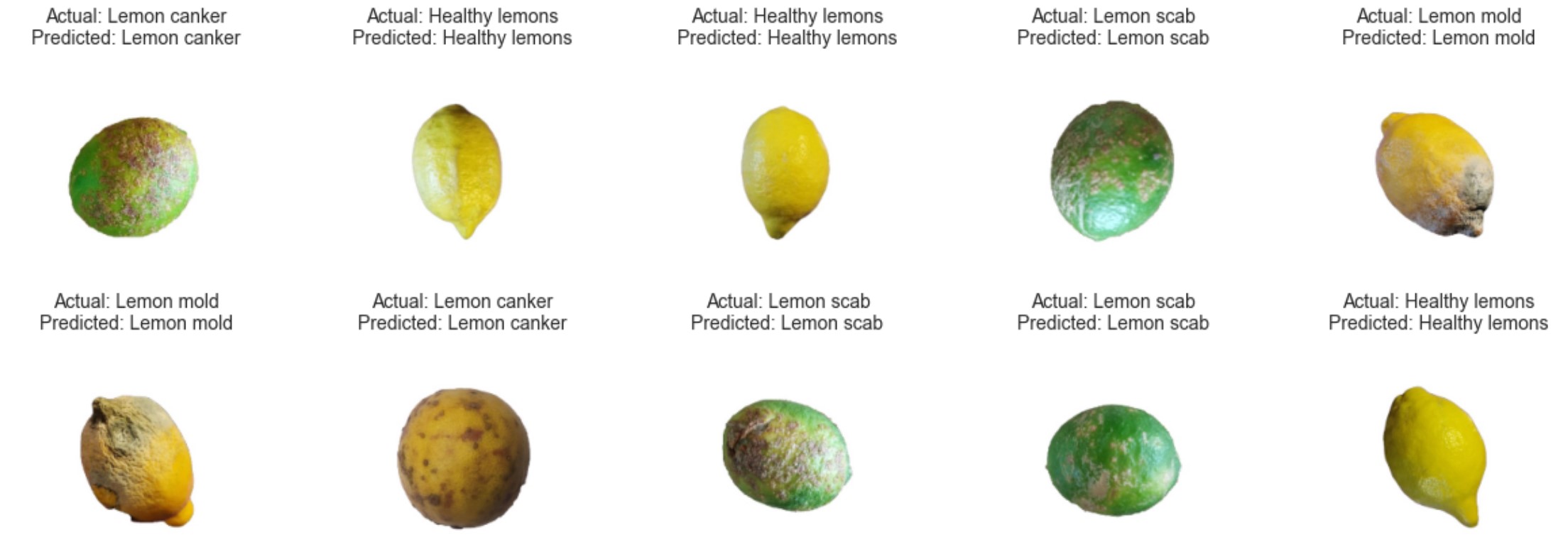}
  \caption{Disease Classification using Logistic Regression (Lemon).}
  \Description{This denotes results of Disease Classification using Logistic Regression (Lemon) }
\end{figure}
% Confusion matrix for lemon disease classification using Logistic Regression

\begin{table}[ht]
\centering
\caption{Confusion matrix for lemon disease classification using Logistic Regression}
\label{tab:lemon_samples}
%\begin{tabular}{|c|c|}
%\setlength{\tabcolsep}{12pt}
\begin{tabular}{lcccc}
\toprule
\textbf{Disease class} & \textbf{Healthy} & \textbf{Canker}& \textbf{Mold} & \textbf{Scab}\\
\midrule
Healthy & 6 & 0 & 0 & 0 \\ 
Canker & 0 & 14 & 0 & 1 \\ 
Mold & 0 & 0 & 9 & 0 \\ 
Scab & 0 & 0 & 0 & 10\\ 
\bottomrule
\end{tabular}
\end{table}

Table 3 represents a confusion matrix for lemon disease classification using Logistic Regression. It's x direction indicates the predicted classes and it's y direction indicates the actual classes. From table 3, it is clearly shown that the correctly classified healthy lemons are 6, canker lemons are 14, mold lemons are 9, and scab lemons are 10. On the other hand, the wrongly classified lemon is 1, whereas the canker lemon is predicted to be a scab lemon.

For the orange dataset using Logistic Regression, we get 99.69\% accuracy, 99.67\% recall, 99.65\% precision as well as 99.66\% f1 score. The following Figure 7 shows the classification of disease using Logistic Regression.

\begin{figure}[ht]
  \centering
  \includegraphics[width=\linewidth]{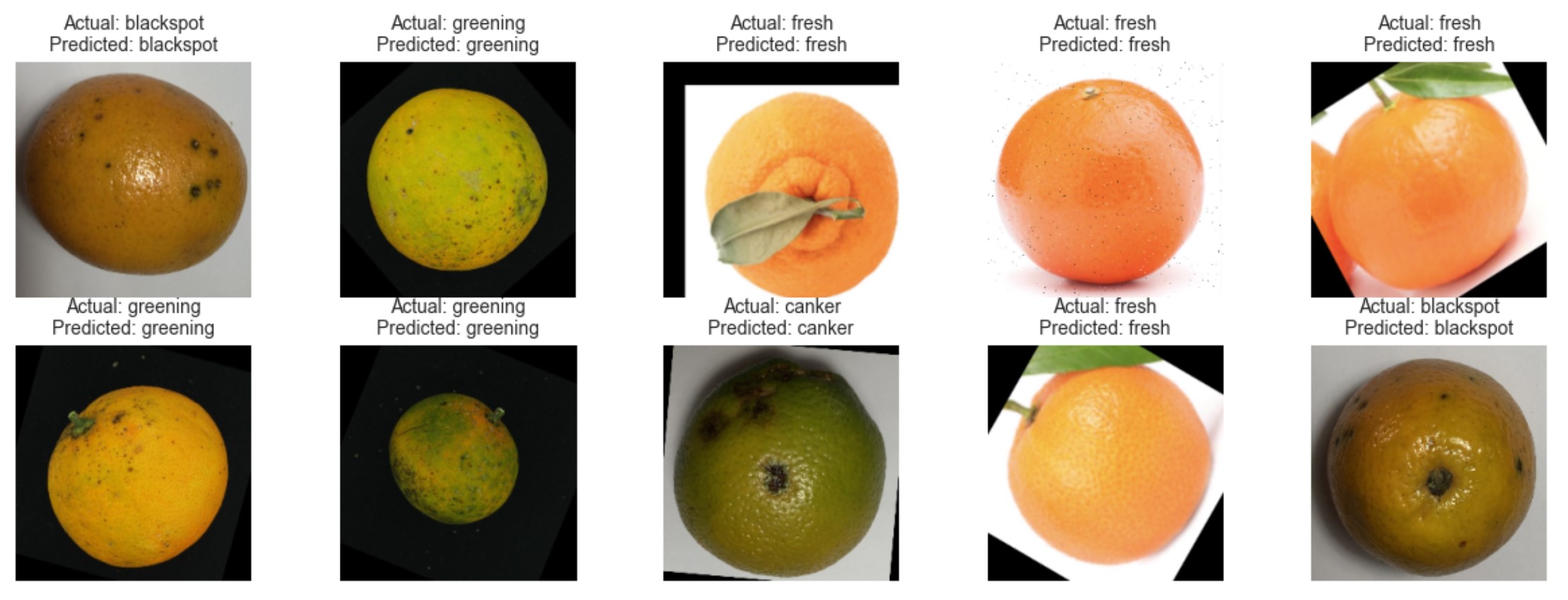}
  \caption{Disease Classification using Logistic Regression (Orange).}
  \Description{This denotes results of Disease Classification using Logistic Regression (Orange)}
\end{figure}

\begin{table}[ht]
\centering
\caption{Confusion matrix for orange disease classification using Logistic Regression}
\label{tab:orange_samples}
%\begin{tabular}{|c|c|}
%\setlength{\tabcolsep}{10pt}
\begin{tabular}{lcccc}
\toprule
%\textbf\centering{Predicted Labels}\\ \hline
\textbf{Disease class} & \textbf{Blackspot} & \textbf{Canker}& \textbf{Fresh} & \textbf{Greening}\\ 
\midrule
Blackspot & 70 & 2 & 0 & 0 \\ 
Canker & 0 & 78 & 0 & 0 \\ 
Fresh & 0 & 0 & 102 & 0 \\ 
Greening & 0 & 0 & 0 & 71\\ 
\bottomrule
\end{tabular}
\end{table}%

Table 4 represents a confusion matrix for orange disease classification using Logistic Regression. It's x direction indicates the predicted classes and it's y direction indicates actual classes. From table 4, it is clearly shown that the correctly classified blackspot oranges are 70, canker oranges are 78, fresh oranges are 102 and greening oranges are 71. 
On the other hand, the wrongly classified blackspot oranges as canker oranges are 2.

%\begin{figure}
%\includegraphics[width=\textwidth]{lrCm (1).jpg}
%\caption{Confusion Matrix for Logistic Regression.} \label{fig6}
%\end{figure}

%\section{Experimental Analysis}
%Finally, Table 5 and Table 6 provide a comparison of the different classifiers along with different pre-trained models for both the lemon and orange datasets. Observing the different performance metrics from the following tables, we can say that Logistic Regression always gives the best results for both datasets when we use ResNet50 as a feature extractor.
%Lemon comparison Table:
\begin{table}[ht]
\centering
\caption{Comparison for Performance Metrics of different Models and Classifiers for Lemon}
\label{tab:model_comparison}
\resizebox{\columnwidth}{!}{
\begin{tabular}{|c|c|c|c|c|c|}
\hline
\textbf{Pre-Trained Model} & \textbf{Classifier} & \textbf{Accuracy} & \textbf{Recall} & \textbf{Precision} & \textbf{F1-score} \\ \hline
VGG16 & KNN & 0.725 & 0.753 & 0.833 & 0.757 \\ \cline{2-6}
 & Random Forest & 0.700 & 0.742 & 0.844 & 0.822 \\ \cline{2-6}
 & Naïve Bayes & 0.800 & 0.789 & 0.806 & 0.787 \\ \cline{2-6}
 & Logistic Regression & 0.850 & 0.866 & 0.870 & 0.850 \\ \hline
VGG19 & KNN & 0.600 & 0.656 & 0.756 & 0.594 \\ \cline{2-6}
 & Random Forest & 0.675 & 0.725 & 0.869 & 0.818 \\ \cline{2-6}
 & Naïve Bayes & 0.800 & 0.786 & 0.807 & 0.775 \\ \cline{2-6}
 & Logistic Regression & 0.875 & 0.894 & 0.892 & 0.888 \\ \hline
\textbf{ResNet50} & KNN & 0.825 & 0.856 & 0.911 & 0.879 \\ \cline{2-6}
 & Random Forest & 0.725 & 0.750 & 0.90 & 0.849 \\ \cline{2-6}
 & Naïve Bayes & 0.825 & 0.858 & 0.856 & 0.856 \\ \cline{2-6}
 
 & \textbf{Logistic Regression} & \textbf{0.950} & \textbf{0.941} & \textbf{0.943} & \textbf{0.938} \\ \hline
 
\end{tabular}
}
\end{table}

%\newpage
%Orange comparison Table:
%\vspace{-10pt}
\begin{table}[ht]
\centering
\caption{Comparison for Performance Metrics of different Models and Classifiers for Orange}
\label{tab:model_comparison1}
\resizebox{\columnwidth}{!}{
\begin{tabular}{|c|c|c|c|c|c|}
\hline
\textbf{Pre-Trained Model} & \textbf{Classifier} & \textbf{Accuracy} & \textbf{Recall} & \textbf{Precision} & \textbf{F1-score} \\ \hline
 VGG16 & KNN & 0.885 & 0.895 & 0.899 & 0.891 \\ \cline{2-6}
 & Random Forest & 0.954 & 0.953 & 0.993 & 0.972 \\ \cline{2-6}
 & Naïve Bayes & 0.873 & 0.873 & 0.874 & 0.872 \\ \cline{2-6}
 & Logistic Regression & 0.994 & 0.993 & 0.993 & 0.993 \\ \hline
  VGG19 & KNN & 0.888 & 0.897 & 0.896 & 0.893 \\ \cline{2-6}
 & Random Forest & 0.916 & 0.915 & 0.996 & 0.953 \\ \cline{2-6}
 & Naïve Bayes & 0.885 & 0.882 & 0.883 & 0.882 \\ \cline{2-6}
 & Logistic Regression & 0.969 & 0.969 & 0.967 & 0.968 \\ \hline
\textbf{ResNet50} & KNN & 0.981 & 0.979 & 0.981 & 0.979 \\ \cline{2-6}
 & Random Forest & 0.981 & 0.979 & 0.991 & 0.985 \\ \cline{2-6}
 & Naïve Bayes & 0.959 & 0.957 & 0.957 & 0.957 \\ \cline{2-6}
  & \textbf{Logistic Regression} & \textbf{0.996} & \textbf{0.996} & \textbf{0.996} & \textbf{0.996}\\ \hline

\end{tabular}
}
\end{table}

Finally, Table 5 and Table 6 provide a comparison of the different classifiers along with different pre-trained models for both the lemon and orange datasets. Observing the different performance metrics from the following tables, we can say that Logistic Regression always gives the best results for both datasets when we use ResNet50 as a feature extractor.

Figures 8 and 9 show the evaluation metrics, which include accuracy, recall, precision, and f1 score, for different classifiers.

\begin{figure}[ht]
  \centering
  \includegraphics[width=\linewidth]{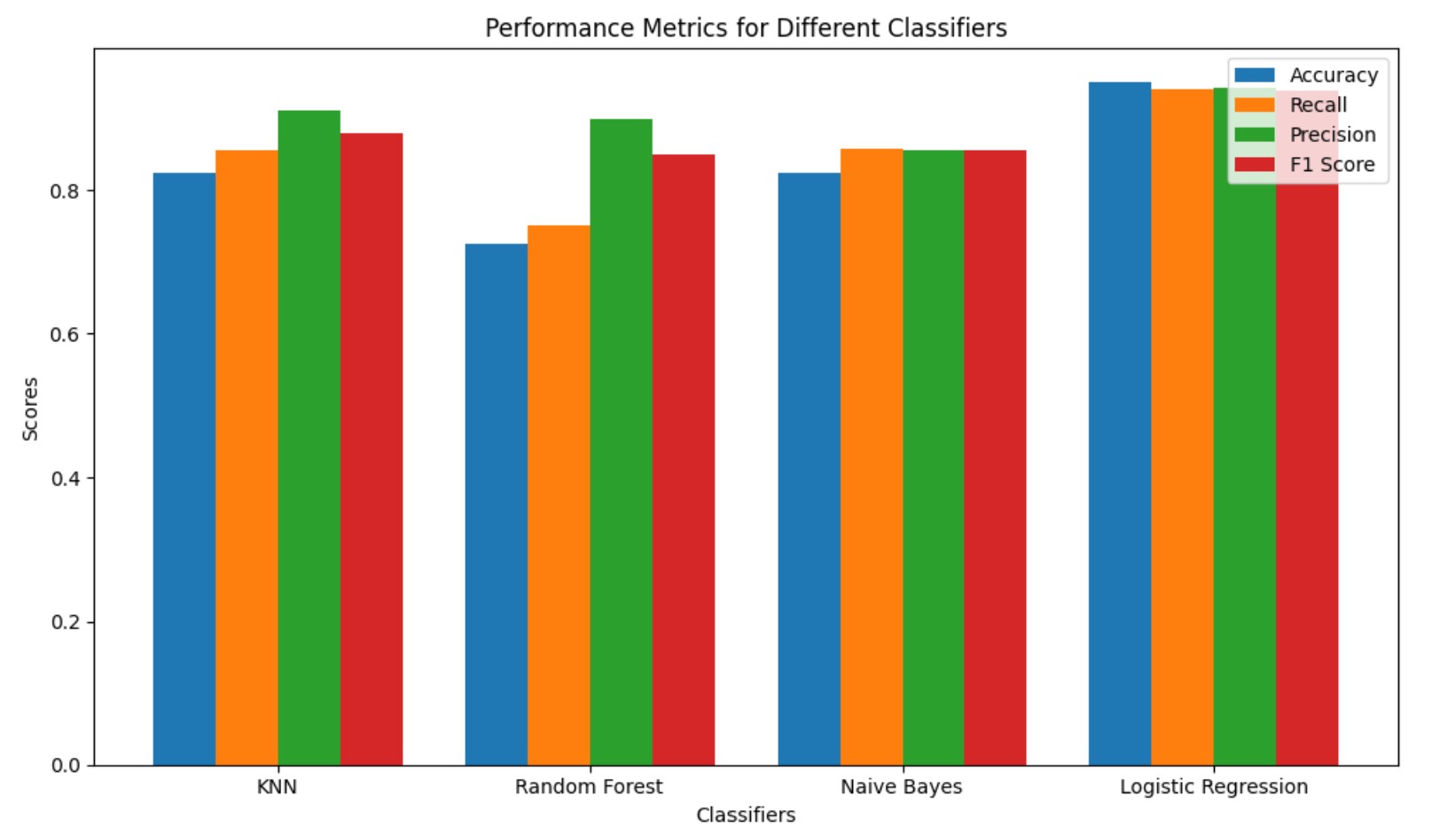}
  \caption{Evaluation Metrics for Various Classifiers (Lemon).}
  \Description{This is evaluation metrics bar chart for lemon}
\end{figure}

\begin{figure}[ht]
  \centering
  \includegraphics[width=\linewidth]{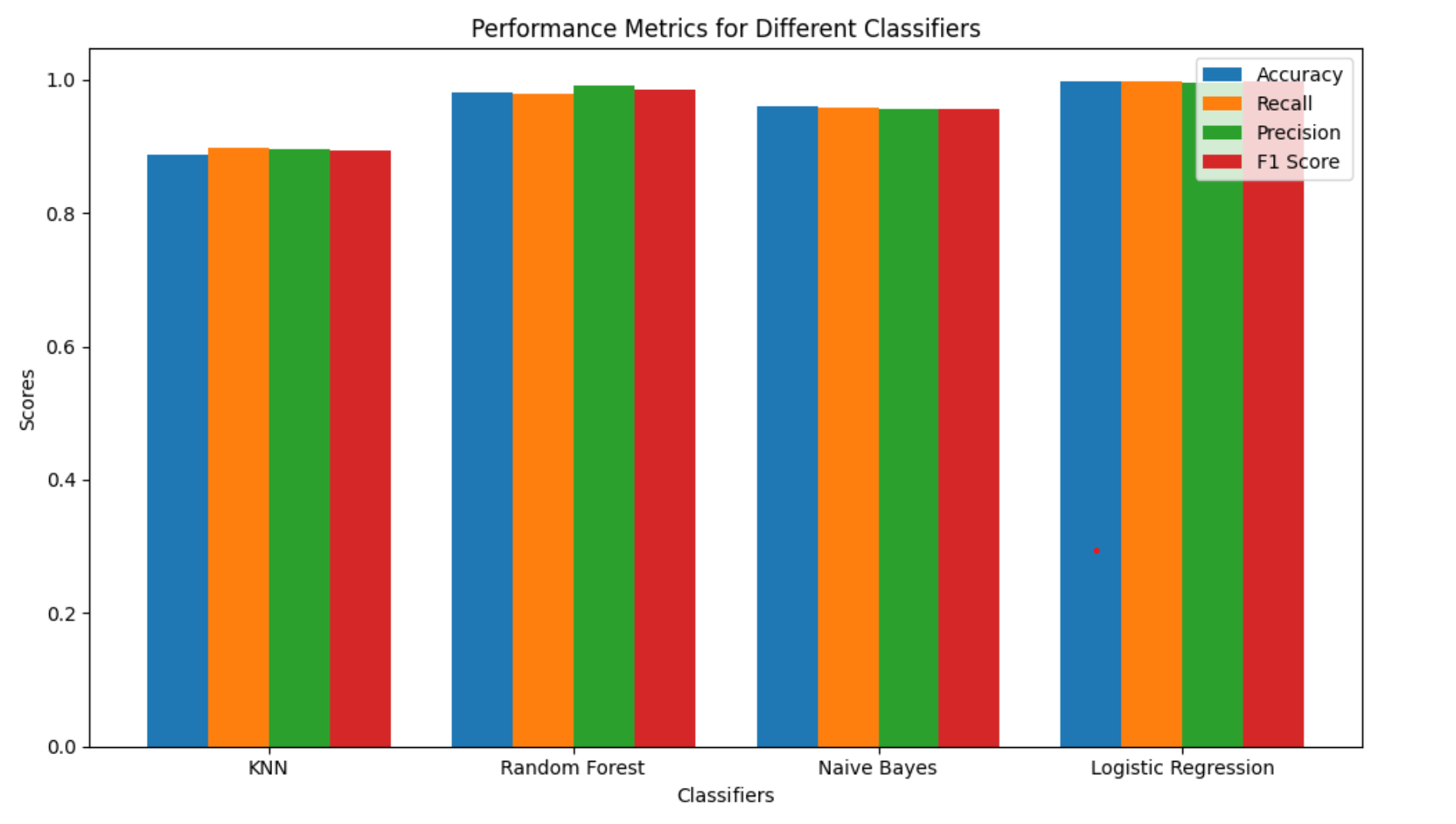}
  \caption{Evaluation Metrics for Various Classifiers (Orange).}
  \Description{This is evaluation metrics bar chart for orange}
\end{figure}

%The evaluation metrics such as accuracy, recall, precision, and f1 score  for various classifiers are visualized in Figures 8 and 9.

 Furthermore, we compared the suggested model's performance to that of two existing models, VGG16 and ResNet50, which are shown in Tables 7 and 8. 
 %These other models directly classify the diseases of lemons and oranges. They didn't use any traditional classification algorithms; they used the CNN model only, specifically VGG16 and ResNet50, as a feature extractor and classifier. We also demonstrate the comparison of the Ensemble method with the existing models in those tables.  

\begin{table}[ht]
\centering
\caption{Performance analysis of different network models for lemon}
\label{tab:orange_samples1}
\resizebox{\columnwidth}{!}{
\begin{tabular}{lcccc}
\toprule
\textbf{Models} & \textbf{Accuracy} & \textbf{Recall} & \textbf{Precision} & \textbf{F1 Score} \\
\midrule
VGG16 & 87.50\% & 87.50\% & 92.36\% & 87.41\% \\ 
ResNet50 & 89.67\% & 88.34\% & 88.43\% & 88.04\% \\ 
%\rowcolor{yellow}
Ensemble & 20.00\% & 20.00\% & 36.33\% & 21.11\% \\ 
ResNet50 with Classifier & 95.00\% & 94.10\% & 94.32\% & 93.88\% \\
\bottomrule
\end{tabular}
}
\end{table}

%\vspace{-25pt}
\begin{table}[ht]
\centering
\caption{Performance analysis of different network models for orange}
\label{tab:orange_samples2}
\resizebox{\columnwidth}{!}{
\begin{tabular}{lcccc}
\toprule
\textbf{Models} & \textbf{Accuracy} & \textbf{Recall} & \textbf{Precision} & \textbf{F1 Score} \\
\midrule
VGG16 & 94.31\% & 94.31\% & 94.40\% & 94.33\% \\ 
ResNet50 & 98.86\% & 98.86\% & 98.87\% & 98.86\% \\ 
%\rowcolor{yellow}
Ensemble & 30.00\% & 30.00\% & 19.00\% & 23.00\% \\ 
ResNet50 with Classifier & 99.69\% & 99.67\% & 99.65\% & 99.66\% \\
\bottomrule
\end{tabular}
}
\end{table}

These other models directly classify the diseases of lemons and oranges. They didn't use any traditional classification algorithms; they used the CNN model only, specifically VGG16 and ResNet50, as a feature extractor and classifier.
%\sethlcolor{yellow}
We also demonstrate the comparison of the Ensemble method with the existing models in those tables.
%\FloatBarrier
\section{Conclusion}
In this study, we successfully classify the diseases of lemons and oranges. In our research, we evaluated various types of CNN models alongside multiple machine learning models and compared their performances. Additionally, we conducted a comparison between two existing models and our proposed models. In this comparative study, our proposed models consistently outperformed the existing ones. When coupled, the ResNet50 and Logistic Regression produced the highest 95\% and 99.69\% accuracy for lemon and orange disease classification, respectively. This results show that deep learning methods can be useful, especially when ResNet50 is combined with more traditional machine learning algorithms like Logistic Regression to accurately classify diseases. This is why we consider this hybrid approach as our proposed model. For the agricultural sector, this model has many advantages. It helps to maintain crop health, which improves yield quantity and quality. 

%\sethlcolor{yellow}
It should be noted that the datasets included in this work have significant constraints due to the fact that all of the images are taken from kaggel. The number of data samples in the lemon dataset is extremely low. In this study, we classified only four classes for both lemon and orange as well as we developed only three CNN architectures.

%\sethlcolor{yellow}
So, our future goal is to collect our own datasets from the real world environment. It is noteworthy that the image quality should be high and the number of images in the datasets need to be increased (mostly for lemon). In future we will try to add more disease classes, and we will try to evaluate our model for other fruit diseases and compare how it performs. Another target is to extend our implementation to assess the disease severity level in lemon and orange fruits. Further, we will implement various CNN-based models for evaluating how models accurately perform.
\bibliographystyle{ACM-Reference-Format}
\bibliography{sample-base.bib}

\appendix

\end{document}